%% file: acl_latex.tex
\pdfoutput=1

\documentclass[11pt]{article}

\usepackage[final]{acl}

\usepackage{times}
\usepackage{latexsym}
\usepackage[T1]{fontenc}

\usepackage[utf8]{inputenc}

\usepackage{microtype}

\usepackage{inconsolata}

\usepackage{mathrsfs}
\usepackage{paralist}
\usepackage{booktabs}
\usepackage{multirow}
\usepackage{pgfplots}
\usepackage{scalefnt}
\usepackage{amsthm,amsmath,amssymb}
\usepackage{mathtools}
\usepackage{bm}
\usepackage{colortbl}
\usepackage{xcolor}
\usepackage{array}
\usepackage{stfloats}
\usepackage{tikz-dependency}
\usepackage{tikz}
\usepackage{siunitx}
\usepackage{graphicx}
\usepackage[ruled]{algorithm2e}
\usepackage{lineno}
\usepackage{subfigure}
\usepackage{helvet}
\usepackage{makecell}
\usepackage{courier}
\usepackage{CJKutf8}
\title{LLMs Can Also Do Well! Breaking Barriers in Semantic Role Labeling via Large Language Models}

\author{
  Xinxin Li$^{1,*}$, Huiyao Chen$^{1,}$\thanks{~~Authors contributed equally.}, Chengjun Liu$^2$, Jing Li$^1$\\
  \textbf{Meishan Zhang$^{1,}$\thanks{~~Corresponding author: Meishan Zhang.}, Jun Yu$^3$, Min Zhang$^1$} \\
  $^1$Institute of Computing and Intelligence, Harbin Institute of Technology (Shenzhen), China\\
  $^2$School of Electrical Engineering and Automation, Harbin Institute of Technology, China\\
  $^3$Computer Science Department, Harbin Institute of Technology (Shenzhen), China\\
  \texttt{\{lixinx0714,chenhy1018,mason.zms\}@gmail.com}\\
}
  
\pgfplotsset{compat=1.18}

\begin{document}
\begin{CJK}{UTF8}{gbsn}
\maketitle
\begin{abstract}
Semantic role labeling (SRL) is a crucial task of natural language processing (NLP).
Although generative decoder-based large language models (LLMs) have achieved remarkable success across various NLP tasks, 
they still lag behind state-of-the-art encoder-decoder (BERT-like) models in SRL.
In this work, we seek to bridge this gap by equipping LLMs for SRL with two mechanisms: (a) retrieval-augmented generation and (b) self-correction.
The first mechanism enables LLMs to leverage external linguistic knowledge such as predicate and argument structure descriptions,
while the second allows LLMs to identify and correct inconsistent SRL outputs.
We conduct extensive experiments on three widely-used benchmarks of SRL (CPB1.0, CoNLL-2009, and CoNLL-2012).
Results demonstrate that our method achieves state-of-the-art performance in both Chinese and English, marking the first successful application of LLMs to surpass encoder-decoder approaches in SRL.
\end{abstract}

\input{sections/1_introduction.tex}
\input{sections/2_related_work.tex}

\input{sections/3_method.tex}
\input{sections/4_exp.tex}
\input{sections/5_conclusion.tex}

\section*{Limitations}
While our proposed framework achieved state-of-the-art performance in SRL, there are several limitations that warrant further exploration.
First, the current self-correction strategy is relatively simple, relying on iterative refinement without explicitly modeling the reasoning process.
This simplicity stems from the need to maintain computational efficiency during large-scale training.
Future work could incorporate advanced techniques, such as chain-of-thought prompting, to enable more structured and interpretable self-correction.
Second, the retrieval-augmented agent currently employs a rule-based traversal approach for candidate predicate retrieval.
This design ensures that the candidate predicates are as comprehensive as possible, providing the LLM with sufficient context for accurate predicate identification.
However, this approach may limit flexibility and scalability.
In the future, we plan to explore the use of generative large language models to dynamically generate candidate predicates, potentially improving both efficiency and accuracy.
Third, the loss function in Equation \ref{eq:final_loss} assigns equal weights to all components, which may not fully capture the varying importance of predicate identification, argument labeling, and self-correction.
Further studies could explore the impact of different weighting strategies on overall performance.
Finally, our approach primarily targets sentence-level SRL, and its scalability to document-level or cross-lingual SRL remains an open challenge, which we aim to address in future research.

\section*{Ethical Statement}
This work does not involve the use of sensitive or private data, and all experiments were conducted on publicly available datasets (CPB1.0, CoNLL-2009, and CoNLL-2012) following their respective usage guidelines.
We have obtained the license for these datasets.
The proposed framework is designed to advance SRL research and does not inherently pose risks of misuse.
However, as with any language model-based system, there is a potential for generating biased or incorrect outputs due to limitations in the training data or model design.
We encourage responsible use of this technology and advocate for ongoing efforts to mitigate bias and ensure fairness in NLP applications.
Additionally, no human subjects were involved in this research, and no ethical concerns were identified during the course of this study.

\section*{Acknowledgements}
We thank the anonymous reviewers for their helpful comments. This work is supported by the National Natural Science Foundation of China (Grant Nos. 62176180), the Shenzhen Science and Technology Program (Grant Nos. ZDSYS20230626091203008 and JCYJ20241202123503005), and the Shenzhen College Stability Support Plan (Grant Nos. GXWD20231130140414001).

\bibliography{custom}

\appendix
\input{sections/6_appendix}
\end{CJK}
\end{document}

%% file: sections/1_introduction.tex
\section{Introduction}

Semantic role labeling (SRL) is a fundamental task in natural language processing \cite[NLP,][]{gildea-jurafsky-2000-automatic}, aiming to analyze the semantic relationships between predicates and their corresponding arguments in a sentence \cite{pradhan-etal-2005-semantic-role,lei-etal-2015-high,chen2025semantic}.
SRL is crucial for various NLP applications, including information extraction \cite{barnickel2009large,christensen-etal-2010-semantic,christensen2011analysis,evans-orasan-2019-sentence}, question answering \cite{shen-lapata-2007-using,berant-etal-2013-semantic,DBLP:conf/acl/YihRMCS16}, machine translation \cite{liu-gildea-2010-semantic,DBLP:conf/acl/ShiLRFLZSW16,marcheggiani-etal-2018-exploiting}, 
robot command parsing \cite{DBLP:conf/acl-jssp/BastianelliCCB13,10.5555/3006652.3006663,thomason2020jointly,DBLP:journals/ftrob/GargSDMCCWCRGCM20,10.1145/3450520} and etc. 
Besides, we can also use SRL to enhance pretrained language models with structured-aware semantic information \cite{DBLP:conf/aaai/0001WZLZZZ20,DBLP:conf/emnlp/XuTSWZSY20}.

\begin{figure}[t]
    \centering
    \includegraphics[width=1.0\columnwidth]{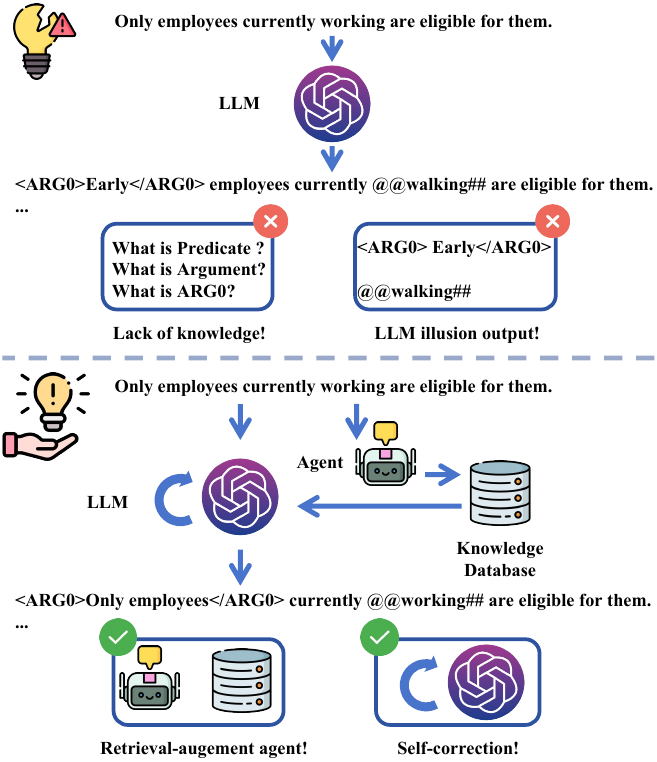}
    \caption{Challenges of direct LLM application in SRL and our solutions. <ARG0> </ARG0> denotes the role, with the enclosed tokens representing the argument, while @@\#\# highlights the predicate.}
    \label{fig:intro}
\end{figure}

Recently, the advent of large language models (LLMs) has reformulated NLP tasks with prompt-based generative solutions \cite{brown2020language}.
Impressive performance has been achieved across a wide range of tasks \cite{achiam2023gpt,wang-etal-2024-knowledgesg,gu2025rapid}.
The reason behind lies in the powerful capability of LLMs for general reasoning \cite{wei2022chain}.
For most NLP tasks, such as information extraction \cite{zhong-etal-2021-factual}, sentiment analysis \cite{radford2019language}, question answering \cite{khashabi-etal-2020-unifiedqa}, and (vallina) machine translation \cite{jiao2023chatgpt}, which require little specialized expertise, LLMs can handle them easily.
Thus, it is reasonable to expect that LLMs can perform well on most of these tasks.

For several NLP tasks, this is not the case.
Unfortunately, SRL is one of them, as it requires strong explainability in linguistics \cite{DBLP:journals/corr/abs-2306-09719,chen-etal-2024-semantic,DBLP:conf/icic/ChengYWLYZTXH24}.
As shown in Figure \ref{fig:intro}, the upper part illustrates a representative baseline of LLM performance in resolving SRL tasks.
Without linguistic knowledge, what is the meaning of predicate-argument structures, \texttt{ARG0$\thicksim$5}, and other related concepts?
Clearly, this is not an easy question that can be answered through general reasoning alone.
A solid background in linguistic knowledge is required to provide an explanation.
Furthermore, the illusion problem in LLMs is another serious issue, as SRL results are highly dependent on the input data.

In this work, we propose two novel mechanisms to sufficiently enhance the expertise of LLMs in addressing the above two problems, respectively.
First, we design a retrieval-augmented agent to enable LLMs to better understand predicates and their semantic arguments.
We construct an external knowledge database based on the predicate-argument description guidance document from the original dataset.
Second, to mitigate the illusion outputs of LLMs, we allow the LLM to verify its outputs by evaluating and correcting them autonomously, a process referred to as self-correction.

Concretely, we design a two-stage conversation-based architecture to perform SRL with LLMs:
(1) predicate identification and (2) argument labeling, which are aligned with traditional BERT-like encoder-decoder systems.
Based on this architecture, we integrate the aforementioned two mechanisms.
LLM reasoning in the two stages is executed in an iterative manner.
In addition, LLM parameters are also optimized to better suit our SRL task.

We conduct comprehensive experiments covering both Chinese (Zh) and English (En) on three widely-used benchmarks: Chinese Proposition Bank 1.0 (CPB1.0) \cite{10.1162/coli.2008.34.2.225}, CoNLL-2009 \cite{hajic-etal-2009-conll}, and CoNLL-2012 \cite{pradhan-etal-2012-conll}.
The results demonstrate that our approach achieves state-of-the-art performance, marking the first time an LLM-driven method has surpassed traditional approaches on the complete SRL task.
This breakthrough highlights not only the effectiveness of leveraging LLMs through our framework but also the critical role of retrieval-augmented agents in addressing the inherent challenges of SRL.

In summary, the main contributions of this work can be summarized as follows:
\begin{compactitem}
    \item We introduce an LLM-driven approach to SRL that annotates predicate-argument triples step-by-step.
    \item We design a retrieval-augmented framework that enhances SRL performance by integrating external knowledge about predicates and their frame descriptions.
    \item We achieve state-of-the-art results on three benchmark datasets, demonstrating the superiority of our approach over traditional methods in complete SRL tasks.
\end{compactitem}
Our code and prompt templates will be publicly available at \href{https://github.com/fangfang123gh/LLM-SRL}{github.com/fangfang123gh/LLM-SRL} to facilitate future research.

%% file: sections/2_related_work.tex
\section{Related Work}
SRL has been explored through transition-based, graph-based, and generative modeling paradigms.

\textbf{Transition-based and Graph-based Methods.}
Transition-based methods construct SRL structures incrementally via a sequence of actions \cite{fernandez-gonzalez-gomez-rodriguez-2020-transition, DBLP:conf/aaai/0001ZLJ21}, while graph-based methods represent predicate-argument structures as graphs, enabling unified modeling of syntactic and semantic dependencies \cite{marcheggiani-titov-2017-encoding, li-etal-2018-unified}. These approaches have achieved strong results, particularly when incorporating syntactic features or external linguistic resources. Recent advances include iterative refinement strategies that progressively improve SRL structures through multiple passes \cite{lyu-etal-2019-semantic}, demonstrating the effectiveness of iterative processing in complex semantic parsing tasks.

\textbf{Generative Approaches.}
Generative methods have shown promising potential by modeling the joint probability distribution of inputs and outputs. Early sequential models like Hidden Markov Models \cite{thompson-etal-2004-generative} laid the foundation, while \citet{yuret-etal-2008-discriminative} captured inter-stage interactions through sophisticated generative frameworks. The emergence of sequence-to-sequence models \cite{daza-frank-2018-sequence} marked a paradigm shift, reformulating SRL as a unified generation task encompassing predicate sense disambiguation, argument identification, and classification. \citet{ijcai2021p521} further advanced this direction with an end-to-end generative framework, achieving excellent performance on both dependency-based and span-based SRL tasks.

\textbf{LLM-based Methods.}
Recently, LLMs have opened new possibilities for SRL through their strong reasoning capabilities. \citet{DBLP:journals/corr/abs-2306-09719} demonstrated the feasibility of using ChatGPT for argument labeling given predicates, while \citet{DBLP:conf/icic/ChengYWLYZTXH24} systematically analyzed the ability of LLMs to capture structured semantics. However, these studies revealed some challenges, including difficulties with predicate-argument structures, absence of explicit domain knowledge, and struggles with self-correction and consistency. These findings underscore the need for integrating external knowledge and iterative reasoning mechanisms to fully realize the potential of LLMs for SRL.

%% file: sections/3_method.tex
\section{Method}
\begin{figure*}[t]
    \centering
    \includegraphics[width=2\columnwidth]{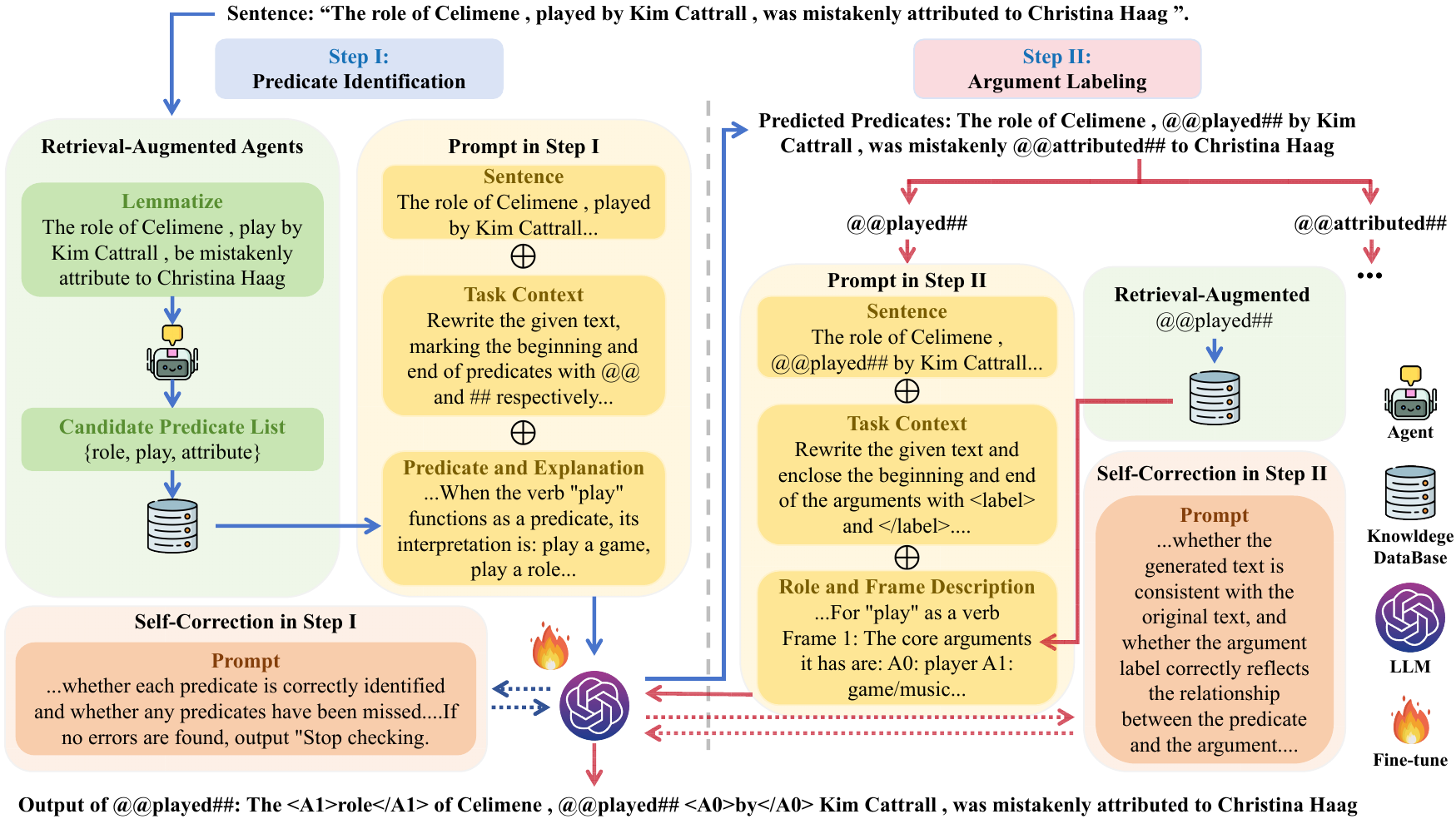}
    \caption{The two-step retrieval-augmented framework for SRL with self-correction mechanism. Step I performs predicate identification with retrieval-augmented prompting, while Step II conducts argument labeling through role and frame descriptions. Both steps incorporate self-correction modules to verify and refine the predictions.}
    \label{fig:framework}
\end{figure*}

In this section, we propose a retrieval-augmented framework for SRL, designed to address challenges such as predicate-argument complexity and the need for external knowledge.
By reformulating SRL as a step-by-step conversational task, the framework integrates retrieval-enhanced agents to incorporate external knowledge and self-correction mechanisms to iteratively refine outputs.
The rationale for unifying the tasks is explained in Appendix \ref{sec:training_strategy_comparison}. 
As illustrated in Figure \ref{fig:framework}, our framework employs a two-stage pipeline where each stage is augmented with retrieval-based knowledge and self-correction mechanisms.
Below, we detail each component along with the mathematical formulations and algorithms involved.
A detailed algorithm description and pseudocode can be found in Appendix~\ref{sec:srl_algorithm}.

\subsection{Preliminary}
Given a sentence input $X = w_1,w_2,\dots,w_n $,
the goal of SRL is to identify all triples $ (P,A,R) = \{(p_1,a_1,r_1), \dots, (p_m,a_m,r_m) \}$,
where $p_i$ represents the predicate,
$a_i$ indicates the associated argument,
and $r_i$ denotes the semantic role assigned to the argument.

\subsection{Predicate Identification}
The first step in SRL is to identify potential predicates within a given sentence.
To address the challenges of predicate recognition, such as ambiguity, we introduce a retrieval-augmented agent that leverages external databases to provide relevant contextual information about predicates.

\paragraph{Special tagging.}
To guide the LLM in predicate identification, we follow \citet{DBLP:journals/corr/abs-2306-09719} by inserting special tags $\text{@@}$ and $\text{\#\#}$ around identified predicates.
This results in sentences annotated with gold-standard predicates:
\begin{equation*}
        Y^{p} = \{w_1, \dots, \text{@@} p_i \text{\#\#}, \dots, w_n\},
\end{equation*}
where for $\forall p \in P$, $\text{@@} p \text{\#\#}$ is an element of $Y^{p}$.

\paragraph{Retrieval-augmented generation.}
To enhance predicate recognition, the retrieval-augmented agent generates a list of candidate predicates and retrieves their corresponding explanations.
Each SRL dataset includes a guideline document with explicit explanations $E_{p_i}$ for each predicate $p_i$.
These documents are organized into a searchable knowledge database.
Concretely, the agent follows a systematic process that begins with lemmatizing words to convert them into their base forms:
\begin{equation*}
        X_{\text{base}} = \text{Lemmatize}(w_1,w_2,\dots,w_n ).
\end{equation*}
Next, the agent iterates through each word in the sentence $X_{\text{base}}$,
constructing a candidate predicate list $\hat{P}=\{ \hat{p}_1, \dots, \hat{p}_k \}$.
For each candidate predicate, the agent retrieves explanations from the knowledge database.
All candidate predicates and their explanations are incorporated as contextual information, culminating in the final prompt $\mathcal{D}^1$:
\begin{equation*}
    \mathcal{D}^1 = X + C^p + \{(\hat{p}_i,E_{\hat{p}_i}) | i = 1,2,\dots,k\},
\end{equation*}
where $C^p$ is the predicate identification task context.
Finally, we feed the prompt $\mathcal{D}^1$ to the LLM to obtain the final output $\widetilde{Y}^{p}$.

\paragraph{Training.}
During training, the cross-entropy objective is used for parameter optimization:
\begin{equation*}
    \mathcal{L}_\text{pred} = - \sum_{t=1}^{|Y^{p}|} \log P(Y_{t}^{p} | Y_{<t}^{p}, \mathcal{D}^1),
\end{equation*}
where $|Y^{p}|$ is the length of the target text $Y^{{p}}$,
and $P(Y_{t}^{p} | Y_{<t}^{p}, \mathcal{D}^1)$ is the probability of generating the word $Y_{t}^{{p}}$ given the input prompt $\mathcal{D}^1$ and the previously generated words $Y_{<t}^{p}$ at time step $t$.

\subsection{Argument Labeling}
After predicate identification, the next step is to assign semantic roles to the arguments associated with the identified predicates.
This involves two subtasks: argument identification and role classification, which are performed simultaneously to minimize error propagation.

\paragraph{Special tagging.}
To identify all arguments related to the given predicate and their corresponding roles,
we insert special tags around each argument,
enclosing the beginning and end with \texttt{<label>} and \texttt{</label>}.
This produces sentences annotated with role tags for all gold-standard arguments associated with the predicate $p_k$:
\begin{equation*}
\begin{split}
Y^{{p}_{k},a,r} &= \{w_1, \dots, \text{<}r_i\text{>} a_i \text{</}r_i\text{>}, \dots, \\
&\quad~~~ \dots, \text{@@} p_k \text{\#\#}, \dots, w_n\}~~~~~~~~~,
\end{split}
\label{eq:arg_representation}
\end{equation*}
where for $\forall (p_k,a,r) \subseteq (P,A,R)$, $\text{<}r\text{>} a \text{</}r\text{>}$ is an element of $Y^{{p}_{k},a,r}$.

\paragraph{Argument-role generation.}
Arguments are divided into:
(1) $\textbf{Core arguments}$ $\mathcal{R}^{\text{core}}$: essential for the predicate, typically labeled as \texttt{ARGN} (\texttt{N} = 0 $\thicksim$ 5), though these labels can be ambiguous without additional context.
(2) $\textbf{Adjunct arguments}$ $\mathcal{R}^{\text{adjunct}}$: Non-essential but supplementary, with consistent labels across predicates.
To refine core argument labels, we use predicate-specific frame descriptions, reducing ambiguity and simplifying the label set.
The complete argument label set $\mathcal{R}_k$ for the specific predicate $p_k$ is:
\begin{equation*}
\mathcal{R}_k = \mathcal{R}^{\text{core}}_{k} \cup \mathcal{R}^{\text{adjunct}},
\end{equation*}
where $\mathcal{R}_k^{\text{core}} \subseteq \mathcal{R}^{\text{core}}$ represents the labels derived from frame descriptions.
This ensures $|\mathcal{R}_k| \leq |\mathcal{R}|$, and $\mathcal{R}$ denotes the overall label set.

To utilize this information, we retrieve all frame descriptions $f_{\text{desc}}$ associated with the base form of the given predicate and include them in the prompt.
This process can be expressed as:
\begin{equation*}
    \mathcal{D}_k^2 = \mathcal{D}^1 + \widetilde{Y}^{{p}} + C^{a} + \mathcal{R}_k + f_{\text{desc}},
\end{equation*}
where, $C^{a}$ represents the argument labeling task context.
The prompt $\mathcal{D}_k^2$ is then fed to the LLM to generate the argument-role result $\widetilde{Y}^{{p}_k,a,r}$ for the specific predicate $p_k$.

\paragraph{Training.}
The cross-entropy objective is used for parameter optimization during training:
\begin{equation*}
    \mathcal{L}_\text{arg} = - \sum_{k=1}^z \sum_{t=1}^{|Y^{{p}_{k},a,r}|} \log P(Y_{t}^{{p}_{k},a,r} | Y_{<t}^{{p}_{k},a,r}, \mathcal{D}_k^2),
\end{equation*}
where $z$ represents the number of gold predicates in $\widetilde{Y}^{{p}}$.

\subsection{Self-Correction Mechanism}
To address challenges such as inconsistency and illusion in LLM-generated outputs, we integrate a self-correction mechanism into our framework.
This mechanism enables the LLM to autonomously identify, evaluate, and refine errors in its predictions through an iterative process.

Once the initial results for either predicate identification or argument labeling are generated, the LLM is prompted to evaluate its outputs for inconsistencies or errors.
If any issues are detected, the LLM adjusts its predictions accordingly.
This process continues iteratively until either the maximum number of iterations $N$ is reached or the model determines that no further corrections are necessary.

The self-correction mechanism for predicate identification is represented as follows:
\begin{equation*}
    \mathcal{D}^1_i =\begin{cases}
        \mathcal{D}^1 + \widetilde{Y}^{{p}} + C_{iter}^p, & \text{if}~~ i = 1 \\
        \mathcal{D}^1_{i-1} + \widetilde{Y}^{{p}}_{i-1} + C_{iter}^p + \widetilde{e}^{{p}}_{i-1}, & \text{else}
        \end{cases}
\end{equation*}
\begin{equation*}
    \widetilde{e}_i^{{p}},\widetilde{Y}_i^{{p}} = \text{LLM}(\mathcal{D}^1_i),
\end{equation*}
where $C_{iter}^p$ represents the context for self-correction in predicate identification,
and $\widetilde{e}_i^{{p}}$ denotes the identified errors during the $i$-th iteration, which are used to refine the predictions.

During training, the gold-standard error $e_i^{{p}}$ for the $i$-th iteration can be directly derived.
Therefore, the gold-standard output for the $i$-th iteration is defined as: $Y^{{p}}_{i} = e_i^{{p}} + Y^{{p}}$.
The self-correction loss for predicate identification is then computed as:
\begin{equation*}
    \mathcal{L}_\text{sc}^\text{pred} = - \sum_{i=1}^{N} \sum_{t=1}^{|Y^{{p}}_{i}|} \log P(Y^{{p}}_{i,t} | Y^{{p}}_{i,<t}, \mathcal{D}^1_i).
\end{equation*}
In the same way, we can also get the self-correction loss for argument labeling task $\mathcal{L}_\text{sc}^\text{arg}$.
Finally, the overall self-correction loss is $\mathcal{L}_\text{sc} = \mathcal{L}_\text{sc}^\text{pred} + \mathcal{L}_\text{sc}^\text{arg}$.

\subsection{Training Loss}
The overall training loss $\mathcal{L}$ for the LLM is defined as a combination of three core components:
the predicate identification loss $\mathcal{L}_{\text{pred}}$,
the argument labeling loss $\mathcal{L}_{\text{arg}}$,
and the self-correction loss $\mathcal{L}_{\text{sc}}$.
This unified objective function is represented as:
\begin{equation}
    \mathcal{L} = \mathcal{L}_{\text{pred}} + \mathcal{L}_{\text{arg}} + \mathcal{L}_{\text{sc}}.
\label{eq:final_loss}
\end{equation}
By jointly optimizing these three components, the framework achieves a balanced and robust learning process, enhancing the overall performance of the model in SRL tasks.

%% file: sections/4_exp.tex
\section{Experiments}
\subsection{Settings}

\paragraph{Dataset.}
\input{tables/data_statistics}
We conduct our experiments on three widely used datasets: CPB1.0 \cite{10.1162/coli.2008.34.2.225} for Chinese, CoNLL09 \cite{hajic-etal-2009-conll} for both English and Chinese, and CoNLL12 \cite{pradhan-etal-2012-conll} for English.
The dataset statistics are summarized in Table~\ref{tab:data_statistics}.
Among them, CPB1.0 and CoNLL12 are span-based SRL datasets, while CoNLL09 is dependency-based.
For CoNLL12, following \citet{he-etal-2018-jointly}, we extract data from OntoNotes \cite{pradhan-etal-2013-towards} and adopt the standard data splits provided by the CoNLL12 shared task \cite{pradhan-etal-2012-conll}.
Given the differences in predicate annotation across datasets, we process them accordingly.
For CPB1.0 (Zh) and CoNLL09 (Zh), where frame files lack explicit predicate explanations, we use the frames as contextual input and employ GPT-4o-mini to generate predicate explanations within the given framework.
To accelerate the selection of the optimal training step checkpoint, we randomly sample 800 sentences from the validation sets of CoNLL09 and CoNLL12 to create a new validation set.
As the validation set of CPB1.0 (Zh) contains fewer than 800 sentences, resampling is not required.

\paragraph{Evaluation metrics.}
Following previous works on SRL \cite{zhou-etal-2020-parsing, zhang-etal-2022-semantic}, we exclude predicate sense disambiguation from our evaluation.
The performance is measured based on atomic predicate-argument structures, represented as tuples in the form of \texttt{<predicate, argument, role>}.
A tuple is deemed correct only if the predicate, argument span boundaries, and role all exactly match the gold-standard annotations.
For span-based SRL datasets, we report precision, recall, and $\text{F}_1$ scores using the official evaluation script.\footnote{\href{https://www.cs.upc.edu/~srlconll/st05/st05.html}{https://www.cs.upc.edu/~srlconll/st05/st05.html}}
For dependency-based SRL datasets, we adopt the official CoNLL-2009 scoring script.\footnote{\href{https://ufal.mff.cuni.cz/conll2009-st/scorer.html}{https://ufal.mff.cuni.cz/conll2009-st/scorer.html}}

\paragraph{Model Details.}
The knowledge databases of each datasets are constructed using frame files provided by the respective datasets.
Specifically, the frame file for CoNLL12 is derived from PropBank\footnote{\href{https://github.com/propbank/propbank-frames}{https://github.com/propbank/propbank-frames}}, which defines the frames of the predicates and provides explanations for their corresponding core arguments.
The retrieval-augmented agent is designed based on rule matching, enabling it to retrieve explanations relevant to the SRL task by leveraging word-based queries.
For word lemmatization, the agent employs the Lemminflect tool\footnote{\href{https://github.com/bjascob/LemmInflect}{https://github.com/bjascob/LemmInflect}}.
The agent's performance on predicate identification is reported in Appendix~\ref{sec:agent_predicate_accuracy}.
We utilize different LLMs for English and Chinese datasets.
For English datasets, we adopt Llama-3-8B-Instruct\footnote{\href{https://huggingface.co/meta-llama/Meta-Llama-3-8B-Instruct}{https://huggingface.co/meta-llama/Meta-Llama-3-8B-Instruct}} \cite{DBLP:journals/corr/abs-2407-21783}, while for Chinese datasets, we use Qwen2.5-7B-Instruct\footnote{\href{https://huggingface.co/Qwen/Qwen2.5-7B-Instruct}{https://huggingface.co/Qwen/Qwen2.5-7B-Instruct}} \cite{DBLP:journals/corr/abs-2412-15115}.
We also evaluated different sizes of Qwen2.5 on CPB1.0, and the detailed results are provided in Appendix~\ref{sec:llm_scale}.
By default, we leverage the LLama-Factory framework\footnote{\href{https://github.com/hiyouga/LLaMA-Factory}{https://github.com/hiyouga/LLaMA-Factory}} \cite{zheng-etal-2024-llamafactory} for parameter updates and fine-tune the LLMs using LoRA \cite{DBLP:conf/iclr/HuSWALWWC22} to enable efficient learning, which is kept at default settings.
The proportion of trainable parameters is provided in the Appendix~\ref{sec:trainable_para}.
Examples of prompts are provided in Appendix~\ref{sec:prompt_template}.

\input{tables/res_2005_2012}
\input{tables/chinese_result}
\paragraph{Hyperparameters.}
All experiments are conducted on a single NVIDIA A800 Tensor Core GPU (80GB).
Each setting is run three times with different random seeds, and the median evaluation scores are reported.
We fine-tuned the LLMs for up to 200,000 steps with a learning rate of 1e-4, saving checkpoints every 20,000 steps.
Best parameters retrieved from development sets are applied to each experiment on test sets.
During training, the number of self-correction iterations $N$ is fixed to 3 in both two steps.

\paragraph{Baselines.}
In this work, we select several recent works as baselines:

\begin{itemize}

    \item \textbf{Traditional methods} are broadly categorized into transition-based \cite{DBLP:journals/kbs/Fernandez-Gonzalez23} and graph-based methods \cite{zhou-etal-2020-parsing, fei-etal-2021-better, zhang-etal-2022-semantic}, both of which have shown outstanding performance.
    Transition-based methods incrementally build SRL structures through a sequence of transition operations, while graph-based methods treat SRL as a graph parsing problem, explicitly modeling predicate-argument relationships.
    \item \citet{DBLP:journals/corr/abs-2306-09719} propose a few-shot SRL method using ChatGPT, assuming predicates are given.
    Sentence embeddings are used to retrieve examples via kNN, and arguments and their corresponding roles are identified by querying ChatGPT for the given predicate.
    \item \citet{DBLP:conf/icic/ChengYWLYZTXH24} design a four-stage SRL pipeline for a given predicate: predicate disambiguation, role retrieval, argument labeling, and post-processing.
    They evaluate the effectiveness of LLMs in a few-shot setting.

\end{itemize}

\subsection{Main Results}

The main results are presented in Table~\ref{tab:result-english} and Table~\ref{tab:result-chinese}.
Our study focuses on generating both predicates and their associated arguments, while prior works \cite{DBLP:journals/corr/abs-2306-09719, DBLP:conf/icic/ChengYWLYZTXH24} typically assume predicates are pre-identified and focus solely on argument recognition.
To ensure a fair comparison, we also report results under the predicate-given setting.

The results reveal that using LLMs for SRL tasks is inherently challenging.
When the LLM is frozen, its $\text{F}_1$ scores range from 2 to 25, significantly lower than traditional methods.
This underscores the difficulty of directly applying LLMs to SRL, particularly in settings without pre-identified predicates.

However, despite these challenges, our retrieval-augmented LLM-based framework consistently outperforms traditional methods in many settings.
For example, in the setting without pre-identified predicates, our method achieves improvements over the previous state-of-the-art by $\bf +0.37~\text{F}_1$ on CoNLL09-WSJ (En), $\bf +0.74~\text{F}_1$ on CoNLL09-Brown (En), and $\bf +0.16~\text{F}_1$ on CoNLL12 (En).
The larger gain on the out-of-domain CoNLL09-Brown (En) test set highlights the strong domain adaptation capabilities of our approach.
Similarly, on Chinese datasets, our method achieves $\bf +2.74~\text{F}_1$ on CPB1.0 (Zh) and $\bf +1.39~\text{F}_1$ on CoNLL09 (Zh) without pre-identified predicates.
\citet{fei-etal-2021-better} incorporate additional syntactic information and achieve superior results on CoNLL09-WSJ (En) and CoNLL12 (En) under the setting where predicates are given. However, without leveraging syntactic information, their reported F1 scores are 92.05 on CoNLL09-WSJ (En) and 85.79 on CoNLL12 (En). In comparison, our approach attains a comparable F1 score on CoNLL09-WSJ (En) and achieves $+2.33~\text{F}_1$ on CoNLL12 (En).
These results demonstrate that our framework effectively addresses the challenges faced by SRL and achieves robust performance in different languages and domains.

When compared with other LLM-based approaches, our method exhibits clear advantages.
Prior works \cite{DBLP:journals/corr/abs-2306-09719, DBLP:conf/icic/ChengYWLYZTXH24} conducted experiments under the predicate-given setting, which simplifies the SRL task.
Even in this setting, our method achieves substantial improvements.
For instance, \citet{DBLP:journals/corr/abs-2306-09719} report $\text{F}_1$ scores of 84.8 and 82.8 on CoNLL09-WSJ (En) and CoNLL12 (En), respectively, while our framework achieves $\bf +7.09~\text{F}_1$ and $\bf +5.32~\text{F}_1$ improvements on these datasets.
These results strongly validate the effectiveness of our two-step retrieval-augmented framework and demonstrate its ability to outperform existing LLM-based methods.

Overall, the results highlight that while SRL remains a challenging task for LLMs, our proposed retrieval-augmented framework not only addresses these challenges but also surpasses both traditional methods and other LLM-based approaches.

\subsection{Analysis}
\paragraph{Ablation experiments.}
\input{tables/ablation}
To assess the contribution of each proposed component, we conducted ablation experiments by removing them individually and evaluating performance on the CPB1.0 (Zh) and CoNLL09-WSJ (En) test sets, as shown in Table~\ref{tab:ablation}.
The results confirm that each component significantly enhances performance, underscoring their importance in achieving optimal results.

Among these, the retrieval-enhanced agent and the role and frame description stand out as key contributors.
The retrieval-enhanced agent leverages external resources to generate a list of candidate predicates along with their explanations, thereby improving predicate recognition accuracy.
This improvement cascades into more accurate argument labeling.
Similarly, the role and frame description aids the LLM in better understanding predicate meanings and role semantics, further boosting overall performance.
In contrast, self-correction had relatively little impact.
As the model adapts to the task with increased training steps, self-correction becomes less critical, especially for simpler sentence annotations.
Removing these components results in performance declines of $\bf 7.92\%$ and $\bf 9.44\%$, respectively, demonstrating the overall effectiveness of our proposed method.

\begin{figure}[t]
    \centering
    \includegraphics[width=1\columnwidth]{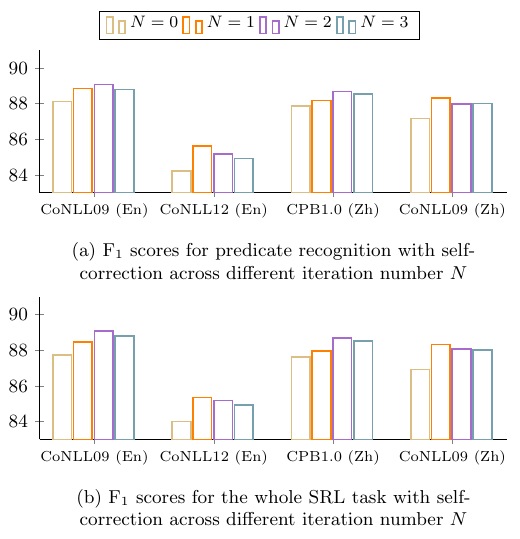} 
    \caption{The results of varying iterations of self-correction on predicate identification and SRL. In (a), the number of iterations for predicate identification varies, while the iterations for argument labeling are fixed at 3. In (b), the number of iterations for the two steps is kept consistent.}
    \label{fig:check_number}
\end{figure}

\paragraph{The impact of iteration $N$.}
The impact of iteration number $N$ on performance is illustrated in Figure~\ref{fig:check_number}, which presents $\text{F}_1$ scores for both predicate identification and the whole SRL task under varying iteration numbers.

The results show that increasing the number of iterations initially improves performance but leads to declines when $N$ becomes too large.
This trend could be attributed to two factors:
(1) the varying difficulty levels across datasets, which require different optimal iteration counts for best performance,
and (2) the potential inaccuracies in error feedback from the LLM, where excessive iterations may amplify deviations from the desired output instead of correcting earlier errors.
Notably, for span-based datasets such as CPB1.0 (Zh) and CoNLL12 (En), the optimal performance is achieved at $N = 1$, while for the dependency-based dataset CoNLL09, the best results are observed at $N = 2$.
Furthermore, the performance trends for the entire SRL task closely mirror those for predicate identification, underscoring the critical role of accurate predicate identification in enhancing the subsequent argument labeling process.

\input{tables/case_study}
\paragraph{A case study of self-correction.}
\label{english_case_study}
Table~\ref{tab:case_study} illustrates the impact of self-correction on predicate identification and argument labeling.
The cases show that self-correction effectively mitigates errors caused by formatting inconsistencies, illusions, and the inherent complexity of the SRL task.

First, self-correction addresses formatting errors and inconsistencies between the generated text and the original.
For instance, in Case 1, the predicate \texttt{"bought"} was missing.
After self-correction, the LLM successfully restores the missing predicate and aligns the output with the intended format.
Second, self-correction resolves errors in role tags.
As shown in Case 3 and Case 4, the LLM initially generates inconsistent role tags \texttt{$<$AD$>$recently$<$/ADV$>$} and incomplete role tags \texttt{PNC$>$to$<$/PNC$>$}, which are corrected through self-correction, resulting in a more coherent and accurate output.
Finally, the iterative nature of self-correction helps refine complex SRL outputs.
For example, in Case 5, the argument was initially labeled as \texttt{"of <A1>going</A1>"}, where the boundary of the argument tag was misplaced.
After self-correction, the LLM correctly identifies and adjusts the argument to \texttt{"<A1>of going</A1>"}, ensuring an accurate representation of the argument's scope.

%% file: tables/data_statistics.tex
\begin{table}[t]
    \centering
    \setlength{\tabcolsep}{10pt}
    \resizebox{1.0\columnwidth}{!}{
        \begin{tabular}{l  cc cc}
            \toprule
            \multicolumn{2}{c}{Dataset}  & \#Sent & \#PA-Triples &\#Roles \\
            \hline

             \multirow{3}{*}{CPB1.0 (Zh)}
             & Train & 8,665 & 65,809 & 20 \\
             & Devel & 549 & 4,445 & 16 \\
             & Test & 983 & 8,001 & 17 \\
             \hline

             \multirow{4}{*}{CoNLL09 (En)}
             & Train & 38,770 &36,5708 &52 \\
             & Devel & 1,334 & 12,918 & 32 \\
             & Re-Devel & 800 &  7,690 & 29 \\
             & Test & 2,396 & 21,634 &36 \\
             & OOD & 421 & 2,766 & 27 \\
            \hline
             \multirow{4}{*}{CoNLL09 (Zh)}
             & Train & 22,276 &231,849 &35 \\
             & Devel &1,762 &18,554 &24 \\
             & Re-Devel & 800 &8,302 &24\\
             & Test & 2,556 &27,712 &26 \\
            \hline
             \multirow{4}{*}{CoNLL12 (En)}
             & Train & 75,187 &566,718 &62 \\
             & Devel & 9,603 &70,871 &46 \\
             & Re-Devel & 800 &5,796 &32 \\
             & Test & 9479 &72879 &49 \\

        \bottomrule
    \end{tabular}
    }
    \caption{Overview of datasets, where \#PA-Triples denotes the number of automatic predicate-argument tuples: \texttt{<predicate, argument, role>}. Meanwhile, \#Sent refers to the number of sentences, and \#Roles indicates the number of roles. }
    \label{tab:data_statistics}
    \end{table}

%% file: tables/res_2005_2012.tex
\begin{table*}[t]
\fontsize{8.5}{9}\selectfont
\setlength{\tabcolsep}{1.3mm}
\begin{center}
\resizebox{1\textwidth}{!}{
\begin{tabular}{lcccccccccccccccccc}
\Xhline{0.08em}
\multirow{4}{*}{\bf Method} & \multicolumn{9}{c}{\bf Without pre-identified predicates} & \multicolumn{9}{c}{\bf With pre-identified predicates}\\
\cmidrule(lr){2-10}\cmidrule(lr){11-19} & \multicolumn{6}{c}{\bf CoNLL09} & \multicolumn{3}{c}{\multirow{2}{*}{\bf CoNLL12}} & \multicolumn{6}{c}{\bf CoNLL09} & \multicolumn{3}{c}{\multirow{2}{*}{\bf CoNLL12}}\\
\cmidrule(lr){2-7}\cmidrule(lr){11-16}
& \multicolumn{3}{c}{\bf WSJ} & \multicolumn{3}{c}{\bf Brown} & & & &  \multicolumn{3}{c}{\bf WSJ} & \multicolumn{3}{c}{\bf Brown} & & &\\
\cmidrule(lr){2-19}
 &  \bf P & \bf R & \bf F$_1$ & \bf P & \bf R & \bf F$_1$ & \bf P & \bf R & \bf F$_1$ & \bf P & \bf R & \bf F$_1$ & \bf P & \bf R & \bf F$_1$ & \bf P & \bf R & \bf F$_1$ \\
\hline
\rowcolor{blue!15}
\multicolumn{19}{c}{\textit{\textbf{Traditional Method}}} \\
\citet{li-etal-2020-high}  &86.16 &85.56 &85.86 &74.65 &73.17 &73.90 & - & - & - &91.60 &88.95 &90.26 &82.6 &78.75 &80.63 & - & - & -\\
   
   \quad + BERT &88.77 &88.62 &88.70 &80.01 &79.80 &79.90 & - & - & - &92.59 &90.98 &91.77 &86.49 &83.80 &85.13  & - & - & - \\

\rowcolor{gray!15}
\citet{zhou-etal-2020-parsing} &84.24 &87.55 &85.86 &76.46 &78.52 &77.47  & - & - & - &85.93 &85.76 &85.84 &76.92 &74.55 &75.72  & - & - & -\\
\rowcolor{gray!15}
\quad + BERT &86.77 &88.49 &87.62 &79.06 &81.67 &80.34 & - & - & - &89.04 &88.79 &88.91 &81.89 &80.98 &81.43 & - & - & - \\

 \citet{zhang-etal-2022-semantic} & - & - & - & - & - & - &79.27 &83.24 &81.21 & - & - & - & - & - & -  &83.02 &84.31 &83.66\\
 \quad + BERT  & - & - & - & - & - & -  &84.53 &86.41 &85.45 & - & - & - & - & - & - &87.52 &87.79 &87.66 \\

 \rowcolor{gray!15}
\citet{fei-etal-2021-better} & - & - & - & - & - & - &- &- &- & 92.24 &92.53 &92.45 & - & - & -  &86.51 &85.92 &86.20\\
 \rowcolor{gray!15}
 \quad + +RoBERTa  & - & - & - & - & - & -  &- &- &- & \textbf{92.89} &\textbf{92.80} &\textbf{92.83} & - & - & - &88.09 &\textbf{88.83} &\textbf{88.59} \\

  \citet{DBLP:journals/kbs/Fernandez-Gonzalez23} &85.90 &88.00 &86.90 &74.40 &76.40 &75.40 & - & - & - &90.20 &90.50 &90.40 &80.60 &80.50 &80.60 & - & - & -\\
 \quad+BERT  &87.20 &\textbf{89.80} &88.50 &79.90 &\textbf{81.80} &80.40 & - & - & - &91.30 &\textbf{91.60} &91.40 &84.50 &84.20 &84.40 & - & - & - \\
   
\hline
\rowcolor{blue!15}
\multicolumn{19}{c}{\textit{\textbf{LLM-based Method}}} \\

\citet{DBLP:journals/corr/abs-2306-09719} &  &  &  &  &  &   &  &  &  &  &  &  &  &  &  &  &  & \\

\quad ChatGPT+SimCSE kNN & - & - & - & - & - & -  & - & - & - & - & - &84.80  & - & - & - & - & - &  82.80 \\
\rowcolor{gray!15}
\citet{DBLP:conf/icic/ChengYWLYZTXH24} &  &  &  &  &  &   &  &  &  &  &  &  &  &  &  &  &  & \\
\rowcolor{gray!15}
\quad Llama2-7B+3-shot & - & - & - & - & - & - & - & - & - & - & - & - & - & - & - &1.83 &10.67 &3.13\\
\hline
Ours &  &  &  &  &  &   &  &  &  &  &  &  &  &  &  &  &  & \\
\quad Llama3-8B (Frozen) &2.63 &1.85 &2.17& 3.14 &2.89 &3.01 &6.29 &7.59 &6.88 &4.19 & 4.71& 4.43& 4.70&6.87& 5.58 &21.91 &16.64 &18.91 \\
\quad Llama3-8B (Fine-tune)&\textbf{89.92} &88.23 &\textbf{89.07} &\textbf{83.36} &79.04 &\textbf{ 81.14} &\textbf{85.64} &85.59	&\textbf{85.61} & 92.78 & 91.02 &91.89 &\textbf{88.41} & \textbf{84.21} & \textbf{86.05} & \textbf{88.19} & 88.06 & 88.12 \\ 
\bottomrule
\end{tabular}
}
\caption{
Main results on the CoNLL09 (En) in-domain (WSJ), out-of-domain (Brown), and CoNLL12 (En) test sets, where Ours experiments have the same settings except that they freeze or fine- tuned the LLM.
}
\label{tab:result-english}
\end{center}
\end{table*}

%% file: tables/chinese_result.tex
\begin{table}[t]
    \centering
    \resizebox{1.0\columnwidth}{!}{
        \begin{tabular}{lcccccc}
        \Xhline{0.08em}
        \multirow{3}{*}{\bf Method} & \multicolumn{6}{c}{\bf Without pre-identified predicates} \\
        \cmidrule(lr){2-7}
        & \multicolumn{3}{c}{\bf CPB1.0} & \multicolumn{3}{c}{\bf CoNLL09}\\
        \cmidrule(lr){2-7}
         &  \bf P & \bf R & \bf F$_1$ & \bf P & \bf R & \bf F$_1$   \\
        \hline
        \rowcolor{blue!15}
        \multicolumn{7}{c}{\textit{\textbf{Traditional Method}}} \\
        \citet{xia-etal-2019-syntax} &- &- &81.73 & - & - & -    \\
        \quad + BERT &- &- &85.57 & - & - & -  \\
         \rowcolor{gray!15}
        \citet{li-etal-2020-high} & - & - & - &83.51 &79.59 &81.50  \\
         \rowcolor{gray!15}
        \quad + BERT  & - & - & - &85.90 &84.90 &85.39\\
        \hline
        \rowcolor{blue!15}
        \multicolumn{7}{c}{\textit{\textbf{LLM-based Method}}} \\
        Ours &  &  &  &  &  &  \\ 
        \quad Qwen2.5-7B (Frozen) & 3.22 &3.59& 3.39& 2.16 &1.67 &1.88 \\
        \quad Qwen2.5-7B (Fine-tune) &\textbf{89.24}	&\textbf{87.40}	&\textbf{88.31} &\textbf{87.56} &\textbf{86.02} &\textbf{86.78}\\
        \hline
        \multirow{3}{*}{\textbf{Method}} & \multicolumn{6}{c}{\bf With pre-identified predicates} \\
        \cmidrule(lr){2-7}
        & \multicolumn{3}{c}{\bf CPB1.0} & \multicolumn{3}{c}{\bf CoNLL09}\\
        \cmidrule(lr){2-7}
         &  \bf P & \bf R & \bf F$_1$ & \bf P & \bf R & \bf F$_1$   \\
        \hline
        \rowcolor{blue!15}
        \multicolumn{7}{c}{\textit{\textbf{Traditional Method}}} \\
        
        \citet{xia-etal-2019-syntax} &84.49 &83.34 &83.91 &84.6 &85.7 &85.1 \\
        \quad + BERT  & - & - & - &89.07 &87.71 &88.38\\

         \rowcolor{gray!15}
        \citet{li-etal-2020-high}  & - & - & - &88.35 &83.82 &86.02\\
         \rowcolor{gray!15}
        \quad + BERT   & - & - & - &89.07 &87.71 &88.38 \\
        
        \hline
        \rowcolor{blue!15}
        \multicolumn{7}{c}{\textit{\textbf{LLM-based Method}}} \\
        Ours &  &  &  &  &  &    \\ 
        \quad Qwen2.5-7B (Frozen)  &9.45 &8.46 &8.93 &4.51 &3.42 &3.89\\
        \quad Qwen2.5-7B (Fine-tune) &\textbf{90.60}	&\textbf{88.18}	&\textbf{89.37} &\textbf{89.56} &\textbf{87.82} &\textbf{88.68}\\
        
        \bottomrule
    \end{tabular}
    }
    \caption{Main results on CPB1.0 (Zh) and CoNLL09 (Zh) test sets.}
    \label{tab:result-chinese}
    \end{table}

%% file: tables/ablation.tex
\begin{table}[!t]
    \centering
    \resizebox{1\columnwidth}{!}{
        \begin{tabular}{l cc }
            \toprule
            \textbf{Method} & \textbf{CPB1.0} &\textbf{CoNLL09} \\    
            \hline
              Ours       &\textbf{88.31}     &\textbf{86.78}   \\
            \quad-w/o retrieval-augmented agent           &      85.92      &        82.86    \\ 
            \quad-w/o role and frame description  &86.43   &84.02   \\
            \quad-w/o self-correction           &        87.05      &        85.47    \\
            \quad-w/o all     &      80.39        &  77.34\\
        \bottomrule
    \end{tabular}
    }
    \caption{ The performance of each component based on the CPB1.0 (Zh) and CoNLL09-WSJ (En) test sets.}
    \label{tab:ablation}
    \end{table}

%% file: tables/case_study.tex
\begin{table}[t]
    \centering
    
\begin{tikzpicture}[node distance = 0cm, auto]
		\def\mytabletwo{
			\renewcommand{\arraystretch}{1.2}
            \resizebox{1.0\columnwidth}{!}{
			\begin{tabular}{cc>{\raggedright\arraybackslash}p{8cm}}
                 \\ \toprule
                 \bf{Num} &  & \bf{Outputs} \\ \hline
                 \bf{1} &\textbf{Before} &That country recently @@ 200,000 tons of sugar. \\
                  &\cellcolor{gray!15}\textbf{After} &\cellcolor{gray!15}That country recently @@bought\#\# 200,000 tons of sugar. \\

                 \bf{2} &\textbf{Before} & That's the biggest @@risk\#\# of all. \\
                  &\cellcolor{gray!15}\textbf{After} & \cellcolor{gray!15}That's the $<$EXT$>$biggest$<$/EXT$>$ @@risk\#\# of all.  \\
                  
                 \bf{3} &\textbf{Before} &That $<$A0$>$that country$<$/A0$>$ $<$AD$>$recently$<$/ADV$>$ @@bought\#\# $<$A1$>$200,000 tons of sugar$<$/A1$>$. \\
                  &\cellcolor{gray!15}\textbf{After} &\cellcolor{gray!15} $<$A0$>$That country$<$/A0$>$ $<$ADV$>$recently$<$/ADV$>$ @@bought\#\# $<$A1$>$200,000 tons of sugar$<$/A1$>$. \\
                  
                 \bf{4} &\textbf{Before} & $<$A0$>$We $<$/A0$>$ @@need\#\# to have PNC$>$to$<$/PNC$>$ spend \$ 5 billion properly. \\
                  &\cellcolor{gray!15}\textbf{After} &\cellcolor{gray!15}$<$A0$>$We $<$/A0$>$ @@need\#\# to have $<$PNC$>$to$<$/PNC$>$ spend \$ 5 billion properly.  \\

                 \bf{5} &\textbf{Before} &$<$A0$>$You$<$/A0$>$ are not @@thinking\#\# of $<$A1$>$going$<$/A1$>$, right? \\
                  &\cellcolor{gray!15}\textbf{After} & \cellcolor{gray!15}$<$A0$>$You$<$/A0$>$ are not @@thinking\#\# $<$A1$>$ of going$<$/A1$>$, right?   \\
                 
            \bottomrule
		\end{tabular}}}

		\newlength\figtwoheight
		\newlength\figtwowidth
		
		\newlength\figtwotableinter
		\setlength\figtwotableinter{\heightof{\mytabletwo}}
		\addtolength{\figtwotableinter}{0.5\figtwoheight}

		\node[inner sep=0pt, node distance = \figtwotableinter] (tab) {\mytabletwo};
		
	\end{tikzpicture}
    \caption{Cases about self-correction impact in English datasets, where \textbf{Before} denotes the output before self-correction and \textbf{After} denotes that after self-correction. Cases about the impact of self-correction on Chinese datasets can be found in the Appendix~\ref{sec:self_correction_chinese}.}
    \label{tab:case_study}
    \end{table}

%% file: sections/5_conclusion.tex
\section{Conclusion}

In this work, we proposed a novel retrieval-augmented framework aimed at bridging the performance gap between large language models (LLMs) and traditional encoder-decoder models in semantic role labeling (SRL).
Through the integration of retrieval-augmented agents, which leverage external knowledge about predicate and argument structures, and a self-correction mechanism for iterative refinement, our method addressed key challenges in SRL, including the complexity of predicate-argument structures, the need for domain-specific linguistic knowledge, and the illusion issues often associated with LLM outputs.
Experiments conducted on three widely-used benchmarks—CPB1.0, CoNLL-2009, and CoNLL-2012—in both English and Chinese demonstrated that our approach achieved state-of-the-art results.
Notably, it marked the first instance of an LLM-based method outperforming traditional approaches across complete SRL tasks.
These results highlight the transformative potential of combining LLMs with structured reasoning and external knowledge for tackling complex linguistic tasks, paving the way for further advancements in semantic understanding.

%% file: sections/6_appendix.tex
\section{Comparison of Unified and Separate Training Strategies}
\label{sec:training_strategy_comparison}

\input{tables/training_strategy_comparison}
To validate our unified LLM-based training approach, we compare it against separate training strategies for predicate identification and argument labeling.
As shown in Table~\ref{tab:training_strategy_comparison}, while separate training achieves marginally higher SRL F$_1$ scores (88.55\% vs. 88.04\%, +0.51\% improvement), the unified approach offers several key advantages:

(1) \textbf{Streamlined pipeline}: Single-stage optimization eliminates the complexity of coordinating separate components.
(2)\textbf{Consistent performance}: Our unified LLM-based method (88.04\%) substantially outperforms traditional unified approaches (86.14\%).
(3) \textbf{Practical deployment}: Unified training simplifies model deployment and maintenance.
Notably, our LLM-based approach demonstrates superior performance over traditional methods in both strategies, with particularly strong gains in predicate identification (97.33\% vs. 94.77\% for separate, 97.15\% vs. 93.80\% for unified).

These results justify our design choice of unified prompt-based formulation, achieving competitive performance while maintaining practical advantages for real-world deployment.

\section{Predicate Identification Performance of Rule-based Agent}
\label{sec:agent_predicate_accuracy}

\input{tables/agent_hit_rate}

Our framework employs a rule-based retrieval agent for predicate identification, which operates through lemmatization and exact string matching against a predefined framework database.
While conceptually simple, this component plays a crucial role in supporting subsequent SRL processing.

Table~\ref{tab:agent_hit_rate} presents the predicate identification performance across multiple standard datasets.
A predicate is considered successfully identified when the agent correctly matches both the position and form of the gold-standard annotation.

The results demonstrate excellent performance across most datasets, with hit rates exceeding 99\% in 6 out of 8 test scenarios.
Key observations include:
(1) \textbf{Perfect accuracy}: CPB1.0 and CoNLL12 achieve 100\% hit rates on both training and test sets, indicating strong coverage for these datasets.
(2) \textbf{Consistent performance}: Most datasets maintain hit rates above 99.8\%, demonstrating the reliability of the rule-based approach for in-domain scenarios.
(3) \textbf{Domain sensitivity}: The Brown shows reduced performance (95.00\%), indicating challenges with domain shift and lexical variability compared to the WSJ (99.30\%).

The performance gap between Brown and WSJ test sets highlights the agent's sensitivity to out-of-domain data, suggesting potential benefits from incorporating LLM-based predicate retrieval for enhanced robustness across diverse domains.
Despite these constraints, the rule-based agent provides reliable predicate identification for most scenarios while maintaining computational efficiency.

\section{Self-correction in Chinese datasets}
\label{sec:self_correction_chinese}
\input{tables/chinese_case_study}

Table~\ref{tab:case_study_chinese} illustrates how self-correction improves predicate identification and argument labeling in Chinese datasets, addressing issues such as formatting errors, inconsistencies, and illusions, as described in Section~\ref{tab:case_study_chinese}.
For example, in Case 1, the predicate "追加" (added) was initially mislabeled as "投资" (investment).
After applying the self-correction mechanism, the model successfully rectified this error, aligning the output with the original text.
Similarly, in Case 4, the model initially failed to include the right boundary of the argument labeled "ADV".
Through self-correction, the model refined the argument's scope, ensuring a more accurate representation.
These examples demonstrate that the self-correction mechanism in Chinese datasets operates similarly to its English counterparts by effectively enhancing the accuracy and coherence of the SRL task.

While the self-correction framework remains consistent across languages, its adaptability allows LLMs to leverage linguistic characteristics unique to each language.
In English datasets (as shown in Table~\ref{tab:case_study}), self-correction often focuses on resolving ambiguities in argument roles and refining argument boundaries.
For instance, English SRL cases frequently involve correcting role tags like "ADV" or handling complex discontinuous spans.
In contrast, Chinese SRL tends to encounter challenges related to predicate disambiguation and ensuring the semantic alignment of predicates and arguments, as Chinese predicates often carry more implicit meanings.
This flexibility highlights the ability of our framework to leverage the learning capabilities of the LLMs to identify and adapt the most critical self-correction directions for optimal performance based on the specific characteristics of each language.

\section{Impact of the LLM scale}
\label{sec:llm_scale}

\input{tables/cpb_model_size}

Table~\ref{tab:cpb_model_size} demonstrates the critical importance of fine-tuning and reveals interesting scaling patterns for SRL tasks.
Our analysis yields several key findings:

\textbf{Frozen models struggle regardless of scale.} Even large-scale models perform poorly without fine-tuning: (1) GPT-4o achieves only 10.05\% F$_1$ score, (2) Qwen2.5-72B (Frozen) reaches 14.80\% F$_1$, likely benefiting from its Chinese optimization that aligns with the CPB1.0 dataset, and (3) DeepSeek-V3-Chat shows the best frozen performance at 21.43\% F$_1$, suggesting more effective built-in strategies for SRL tasks.

\textbf{Fine-tuning enables dramatic performance gains.} The transition from frozen to fine-tuned models shows remarkable improvement, with Qwen2.5-1.5B jumping to 83.12\% F$_1$ score. This 69-point improvement underscores that task-specific adaptation is essential for specialized tasks like SRL, regardless of the model's general capabilities.

\textbf{Scaling benefits emerge with fine-tuning.} Among fine-tuned models, larger sizes consistently improve performance: Qwen2.5-3B (84.63\% F$_1$) $\rightarrow$ Qwen2.5-7B (88.31\% F$_1$) $\rightarrow$ Qwen2.5-14B (88.62\% F$_1$). However, the marginal gain diminishes significantly beyond 7B parameters (only 0.31\% improvement from 7B to 14B), suggesting diminishing returns at larger scales.

\textbf{Implications for model selection.} The results indicate that (1) fine-tuning is non-negotiable for SRL tasks, as even 72B frozen models underperform 1.5B fine-tuned ones, (2) moderate scaling (1.5B $\rightarrow$ 7B) provides substantial benefits, but (3) further scaling beyond 7B offers limited gains, emphasizing the need for balanced strategies that combine appropriate model size with effective task-specific adaptations rather than relying solely on parameter scaling.

\section{Discussion on Trainable Parameters}
\label{sec:trainable_para}

\input{tables/model_params}

While LLM-based approaches utilize models with substantially larger total parameter counts, they achieve remarkable parameter efficiency during fine-tuning through parameter-efficient adaptation techniques such as LoRA.

Table~\ref{tab:model_params} presents a comprehensive comparison between established BERT-based SRL methods and our LLM-based approach across key efficiency metrics.
The results reveal a striking contrast in training requirements:
(1) \textbf{BERT-based models require full fine-tuning}: Traditional approaches fine-tune 100\% of their parameters, ranging from 338M to over 510M trainable parameters across different implementations.
(2) \textbf{LLM-based models achieve extreme parameter efficiency}: Our method fine-tunes only 20-21M parameters (approximately 0.26\% of total model size), representing a 15-25× reduction in trainable parameters compared to BERT-based approaches.
(3) \textbf{Efficiency scales favorably}: Despite using 7-8B parameter base models, our approach requires fewer trainable parameters than any BERT-based method, demonstrating superior training efficiency.

We will explore more efficient methods in future work. For now, our primary goal is to establish a strong baseline and bridge the SRL task into the decoder-only LLM paradigm. This direction aligns with the trend of leveraging LLMs as a unified solution for various language understanding tasks.

\section{Description of Our SRL Algorithm}
\label{sec:srl_algorithm}
To further clarify the implementation details of our proposed framework, we provide a detailed explanation of the retrieval-augmented SRL with self-correction algorithm, as outlined in Algorithm \ref{tab:algorithm}.
This algorithm systematically integrates retrieval-augmented agents and self-correction mechanisms into the SRL process, ensuring both accuracy and consistency in predicate-argument-role identification.
Below, we describe the key stages of the algorithm:

\paragraph{Predicate identification with retrieval.}
The process begins by lemmatizing the input sentence to normalize its tokens.
A retrieval-augmented agent then generates a list of candidate predicates by analyzing the lemmatized sentence and retrieving relevant contextual information from a knowledge database.
For each candidate predicate, the agent retrieves corresponding explanations, which are combined with the input sentence and task-specific context to form a prompt.
This prompt is fed into the LLM to identify predicates.

\paragraph{Predicate self-correction.}
After the initial predicate identification, a self-correction mechanism iteratively refines the predictions.
At each iteration, the LLM evaluates its outputs, identifies potential errors, and updates its predictions accordingly.
This process continues until no further errors are detected or the maximum number of iterations is reached.

\paragraph{Argument labeling.}
For each identified predicate, the algorithm retrieves role sets and frame descriptions from the knowledge database.
These are used as contextual prompts to guide the LLM in labeling arguments and assigning semantic roles.

\paragraph{Argument self-correction.}
Similar to predicate self-correction, the algorithm applies an iterative self-correction mechanism to refine argument labeling results.
The LLM evaluates its outputs for consistency and correctness, making adjustments as needed.
\input{tables/algorithm}

\section{Prompts of Our SRL Framework}
\label{sec:prompt_template}
\definecolor{process_text}{RGB}{180,0,3}

\definecolor{agent_text}{RGB}{150,10,200}
\definecolor{llm_text}{RGB}{55,180,9}
This section provides the specific prompts used in our approach to ensure reproducibility.
Detailed prompts for the Chinese dataset will be included in our code repository.

In the prompt templates:
(1) fixed prompts are displayed in black.
(2) Input text is highlighted in \textcolor{process_text}{deep red}.
(3) The candidate predicate list, along with explanations retrieved by the retrieval-augmented agent, is shown in \textcolor{agent_text}{purple}.
(4) The output generated by the LLM is presented in \textcolor{llm_text}{green}.

These conventions are designed to make the prompts clear and easy to follow.

\input{tables/pred_prompt}
\input{tables/pred_self_correction}
\input{tables/arg_prompt}
\input{tables/arg_self_correction}
\input{tables/gpt_template}

%% file: tables/training_strategy_comparison.tex
\begin{table}[h]
    \centering
    \resizebox{1\columnwidth}{!}{
        \begin{tabular}{l ccc }
            \toprule
            \textbf{Training Strategy} & \textbf{Method} & \textbf{Predicate F$_1$ (\%)} & \textbf{SRL F$_1$ (\%)} \\    
            \hline
              Separate Predicate & Traditional & 94.77 & - \\
Separate Argument  & Traditional & 100.00 & 85.80 \\
Unified            & Traditional & 93.80 & 86.14 \\
Separate Predicate & Ours        & 97.33 & - \\
Separate Argument  & Ours        & 100.00 & 88.55 \\
Unified            & Ours        & 97.15 & 88.04 \\

        \bottomrule
    \end{tabular}
    }
    \caption{Performance comparison of separate vs. unified training strategies across traditional and LLM-based approaches. "Separate" denotes independent training of predicate identification and argument labeling, while "Unified" refers to joint end-to-end optimization.}
    \label{tab:training_strategy_comparison}
    \end{table}

%% file: tables/agent_hit_rate.tex
\begin{table}[h]
    \centering
    \resizebox{1\columnwidth}{!}{
        \begin{tabular}{l ll lc }
            \toprule
            \textbf{Dataset} & \textbf{Split} & \textbf{\#Predicates} & \textbf{\#Missed} & \textbf{Hit Rate (\%)} \\    
            \hline
             CPB1.0         & Train    & 30,220  & 1   & 100.00 \\
               & Test     & 3,513   & 0   & 100.00 \\
CoNLL09 (En)   & Train    & 177,971 & 33  & 99.98  \\
               & WSJ Test & 10,498  & 73  & 99.30  \\
               & Brown Test & 1,259 & 63  & 95.00  \\
CoNLL09 (Zh)   & Train    & 102,803 & 124 & 99.88  \\
               & Test     & 12,282  & 20  & 99.84  \\
CoNLL12        & Train    & 75,187  & 0   & 100.00 \\
               & Test     & 9,479   & 0   & 100.00 \\

        \bottomrule
    \end{tabular}
    }
    \caption{Predicate identification performance of the rule-based retrieval agent across different datasets. \#Predicates and \#Missed indicate the total number of predicates and failed retrievals, respectively.}
    \label{tab:agent_hit_rate}
    \end{table}

%% file: tables/chinese_case_study.tex
\begin{table}[ht]
    \centering

\begin{tikzpicture}[node distance = 0cm, auto]
		\def\mytabletwo{
			\renewcommand{\arraystretch}{1.2}
            \resizebox{1.0\columnwidth}{!}{
			\begin{tabular}{cc>{\raggedright\arraybackslash}p{8cm}}
                 \\ \toprule
                 \bf{Num}& & \bf{Outputs} \\ \hline
                 \textbf{1}& \textbf{Before} & 集团在上海追加 @@投资\#\# 二千四百万美元。\\
                 & &The group in Shanghai added an @@investment\#\# of 24 million dollars. \\
                 & \cellcolor{gray!15}\textbf{After} &\cellcolor{gray!15}集团在上海 @@追加\#\# 投资二千四百万美元。  \\

                 & \cellcolor{gray!15}& \cellcolor{gray!15} 
                The group in Shanghai @@added\#\# an investment of 24 million dollars. \\
                 \textbf{2}& \textbf{Before} & $<$A0$>$她$<$/A0$>$ @@说\#\#，$<$A1$>$“化工重点项目建设进展较快。$<$/A1$>$ \\
                 & &

              $<$A0$>$She$<$/A0$>$ @@said\#\#, $<$A1$>$``The construction progress of key chemical engineering projects is relatively fast.$<$/A1$>$ \\
                & \cellcolor{gray!15}\textbf{After} &\cellcolor{gray!15}$<$A0$>$她$<$/A0$>$ @@说\#\#，$<$A1$>$“化工重点项目建设进展较快。”$<$/A1$>$ \\
                & \cellcolor{gray!15} & 

                \cellcolor{gray!15}$<$A0$>$She$<$/A0$>$ @@said\#\#, $<$A1$>$``The construction progress of key chemical engineering projects is relatively fast.''$<$/A1$>$ \\
                  
                 \textbf{3}& \textbf{Before} & 城建成为 $<$A0$>$外商$<$/A0$> $@@投资\#\# 青海新热点。 \\
        
                 & & Urban construction has become a new hotspot for $<$A0$>$ foreign investors $<$/A0$>$ to @@invest\#\#  Qinghai.
                 \\
                 & \cellcolor{gray!15}\textbf{After} & \cellcolor{gray!15}城建成为 $<$A0$>$外商$<$/A0$> $@@投资\#\# $<$A1$>$青海$<$/A1$>$ 新热点。  \\
                 & \cellcolor{gray!15} & \cellcolor{gray!15} Urban construction has become a new hotspot for $<$A0$>$ foreign investors $<$/A0$>$ to @@invest\#\#  $<$A1$>$Qinghai$<$/A1$>$.
                 \\
                  
                 \textbf{4}& \textbf{Before} &$<$A1$>$东亚经济$<$A1$>$ $<$ADV$>$一定$<$/ADV$>$ 能够 $<$ADV$>$继续向前$<$/ADV$>$ @@发展\#\#。 \\
                 & & $<$A1$>$The East Asian economy$<$/A1$>$ $<$ADV$>$will certainly$<$/ADV$>$ able to $<$ADV$>$continue moving forward $<$/ADV$>$ @@develop\#\#.\\
                 & \cellcolor{gray!15}\textbf{After} &\cellcolor{gray!15}$<$A1$>$东亚经济$<$A1$>$ $<$ADV$>$一定$<$/ADV$>$ 能够继续 $<$ADV$>$向前$<$/ADV$>$ @@发展\#\#。 \\
                 & \cellcolor{gray!15} & \cellcolor{gray!15}$<$A1$>$The East Asian economy$<$/A1$>$ $<$ADV$>$will certainly$<$/ADV$>$ able to continue $<$ADV$>$ moving forward $<$/ADV$>$ @@develop\#\#.\\
                 
            \bottomrule
		\end{tabular}}}

		\setlength\figtwotableinter{\heightof{\mytabletwo}}
		\addtolength{\figtwotableinter}{0.5\figtwoheight}

		\node[inner sep=0pt, node distance = \figtwotableinter] (tab) {\mytabletwo};
		
	\end{tikzpicture}
    \caption{Cases about self-correction impact in Chinese datasets, where \textbf{Before} denotes the output before self-correction and \textbf{After} denotes that after self-correction. Each Chinese case is followed by its corresponding English translation.}
    \label{tab:case_study_chinese}
    \end{table}

%% file: tables/cpb_model_size.tex
\begin{table}[h]
    \centering
    \resizebox{1\columnwidth}{!}{
        \begin{tabular}{l ccc }
            \toprule
            \textbf{Method} & \textbf{P} & \textbf{R} & $\textbf{F}_1$ \\    
            \hline
              Ours       &  &  &    \\
              \quad - GPT-4o &8.28 &12.79 &10.05 \\
              \quad - DeepSeek-V3-Chat &14.72 &39.37 &21.43 \\
              \quad - Qwen2.5-72B (Frozen) &13.23 & 16.79 &14.80 \\
              \quad - Qwen2.5-1.5B (Fine-tune) &83.89& 82.36& 83.12\\
              \quad - Qwen2.5--3B (Fine-tune) &85.87 &83.20 &84.63 \\
              \quad - Qwen2.5-7B (Fine-tune) &89.24 &87.40	&88.31 \\
              \quad - Qwen2.5-14B (Fine-tune) &\textbf{89.72}	&\textbf{87.55}	&\textbf{88.62} \\

        \bottomrule
    \end{tabular}
    }
    \caption{Performance comparison of LLMs with varying parameter sizes on CPB1.0 dataset without predicate pre-identification. Results show Precision (P), Recall (R), and F$_1$ for both frozen and fine-tuned models.}
    \label{tab:cpb_model_size}
    \end{table}

%% file: tables/model_params.tex
\begin{table}[h]
    \centering
    \resizebox{1\columnwidth}{!}{
        \begin{tabular}{l llc }
            \toprule
            \textbf{Method} & \textbf{Model Type} & \textbf{\#Trainable} & \textbf{Total} \\    
            \hline
              \citet{li-etal-2020-high} & BERT-large & $>$355M (100\%) & $>$355M \\
\citet{zhou-etal-2020-parsing} & BERT-large & 510.7M (100\%) & 510.7M \\
\citet{zhang-etal-2022-semantic} & BERT-large & 338.6M (100\%) & 338.6M \\
\citet{DBLP:journals/kbs/Fernandez-Gonzalez23} & BERT-large & $>$360M (100\%) & $>$360M \\
\textbf{Ours} & Qwen2.5-7B & \textbf{20.2M (0.26\%)} & 7.6B \\
\textbf{Ours} & LLaMA3-8B & \textbf{21.0M (0.26\%)} & 8.1B \\

        \bottomrule
    \end{tabular}
    }
    \caption{Parameter efficiency comparison between BERT-based and LLM-based SRL models. \#Trainable shows trainable parameters during fine-tuning, with percentages indicating the proportion of total model parameters requiring updates.}
    \label{tab:model_params}
    \end{table}

%% file: tables/algorithm.tex
\begin{algorithm*}[t]
    \caption{Retrieval-Augmented SRL with Self-Correction}
    \label{tab:algorithm}
    \LinesNumbered
    
    \KwIn{
        Input sentence: $X = w_1, w_2, ..., w_n$; Knowledge database: $K_D$; Maximum correction iterations: $N$; Task-specific contexts: $C$ ($C^p$: Predicate identification context, $C_{iter}^p$: Predicate self-correction context, $C^a$: Argument labeling context, $C_{iter}^a$: Argument self-correction context)
    }
    \KwOut{
        SRL triples $(P, A, R) = \{(p_1, a_1, r_1), ..., (p_m, a_m, r_m)\}$
    }
    \tcc{Stage 1: Predicate Identification with Retrieval}
    $X_{base} \gets \text{Lemmatize}(w_1, w_2, ..., w_n)$\;
    Generate candidate predicates $\hat{P} = \{\hat{p}_1, ..., \hat{p}_k\}$ from $X_{base}$ by the retrieval-argumented agent\;
    Retrieve explanations $E_{\hat{p}_i}$ for each $\hat{p}_i$ from $K_D$\;
    $\mathcal{D}^1 \gets X + C^p + \{(\hat{p}_i,E_{\hat{p}_i})|i = 1, 2, ..., k\}$\;
    Generate initial predicate results: $\widetilde{Y}^p \gets \text{LLM}(\mathcal{D}^1)$\;

    \tcc{Predicate Self-Correction}
    \For{$i \gets 1$ \KwTo $N$}{
        \eIf{$i = 1$}{
            $\mathcal{D}_i^1 \gets \mathcal{D}^1 + \widetilde{Y}^p + C_{iter}^p$\;
        }{
            $\mathcal{D}_i^1 \gets \mathcal{D}^1_{i-1} + \widetilde{Y}^p_{i-1} + C_{iter}^p + \widetilde{e}^p_{i-1}$\;
        }
        Generate errors and updated results: $\widetilde{e}^p_{i}, \widetilde{Y}^p_{i} \gets \text{LLM}(\mathcal{D}_i^1)$\;
        \If{no errors detected}{
            break\;
        }
    }

    \tcc{Stage 2: Argument Labeling}
    \ForEach{predicate $p_k$ in final $\widetilde{Y}^p$}{
        Get role set $\mathcal{R}_k = \mathcal{R}^{\text{core}}_{k} \cup \mathcal{R}^{\text{adjunct}}$\;
        Retrieve frame descriptions $f_{desc}$ for $p_k$ from $K_D$\;
        $\mathcal{D}_k^2 \gets \mathcal{D}^1 + \widetilde{Y}^{\text{p}} + C^{a} + \mathcal{R}_k + f_{\text{desc}}$\;
        Generate initial argument results: $\widetilde{Y}^{\text{p}_k,a,r} \gets \text{LLM}(\mathcal{D}_k^2)$\;
        
        \tcc{Self-Correction Arguments for Current Predicate}
        \For{$i \gets 1$ \KwTo $N$}{
            \eIf{$i = 1$}{
                $\mathcal{D}_{k,i}^2 \gets \mathcal{D}_k^2 + \widetilde{Y}^{p_k,a,r} + C_{iter}^a$\;
            }{
                $\mathcal{D}_{k,i}^2 \gets \mathcal{D}^2_{k,i-1} + \widetilde{Y}^{p_k,a,r}_{i-1} + C_{iter}^a + \widetilde{e}^{p_k,a,r}_{i-1}$\;
            }
            Generate errors and updated results: $\widetilde{e}^{p_k,a,r}_{i}, \widetilde{Y}^{p_k,a,r}_{i} \gets \text{LLM}(\mathcal{D}_{k,i}^2)$\;
            \If{no errors detected}{
                break\;
            }
        }
    }
    \Return{SRL triples from all final $\widetilde{Y}^{p_k,a,r}$}

\end{algorithm*}

%% file: tables/pred_prompt.tex
\begin{table}[h]
    \centering
    
\definecolor{process_text}{RGB}{180,0,3}

\definecolor{agent_text}{RGB}{150,10,200}
\definecolor{llm_text}{RGB}{55,180,9}

\begin{tikzpicture}

    \node[inner sep=0pt] (table) {
        \begin{tabular}{p{0.45\textwidth}}
             \rowcolor{gray!15}
            \textbf{Prompt $\mathcal D^{1}$: Predicate Identification }
              \\
            \hline 
            \textbf{System}: You are a helpful assistant who has a background in linguistics and is good at understanding texts, especially skilled in semantic role labeling recognition.
            \\
            \\
            \textbf{User}: Semantic Role Labeling (SRL) aims to identify predicates in a sentence and assign roles to their arguments. A predicate refers to the core word or phrase in a sentence that conveys an action, event, or state and serves as the focus for other elements in the sentence.\\
            \textbf{Text}: \textcolor{process_text}{What was the , purpose and goal of this campaign ?}\\
            For the predicate indentification task, what are the predicates in the given text? \\
            Possible predicate results in the text are: 
            \textcolor{agent_text}{\texttt{"}What @@was\#\# the , @@purpose\#\# and goal of this @@campaign\#\# ?\texttt{"}, where predicates are specified by @@ and \#\#.} \\
            \textcolor{agent_text}{
            When the verb \texttt{"}purpose\texttt{"} functions as a predicate, its interpretation is: purpose.} \\
            \textcolor{agent_text}{
            When the verb \texttt{"}be\texttt{"} functions as a predicate, its interpretation is: copula, existential, auxiliary, be like: multiword expression akin to \texttt{"}say\texttt{"}.} \\
            \textcolor{agent_text}{
            When the verb \texttt{"}campaign\texttt{"} functions as a predicate, its interpretation is: work towards a goal.} \\
            Based on the given possible predicate results and interpretations, rewrite the given text, marking the beginning and end of predicates with @@ and \#\# respectively. Note that words not present in the predicate results may also be predicates.
            \\
            \textbf{LLM}: \textcolor{llm_text}{Predicate identification result.}
            
        \end{tabular}
    };
    
    \draw[rounded corners=5pt, line width=1pt] (table.north west) -- (table.north east) -- (table.south east) -- (table.south west) -- cycle;
\end{tikzpicture}
\end{table}

%% file: tables/pred_self_correction.tex
\begin{table}[ht]
    \centering

\definecolor{process_text}{RGB}{180,0,3}

\definecolor{agent_text}{RGB}{150,10,200}
\definecolor{llm_text}{RGB}{55,180,9}

\begin{tikzpicture}

    \node[inner sep=0pt] (table) {
        \begin{tabular}{p{0.45\textwidth}}
             \rowcolor{gray!15}
            \textbf{Prompt $C^p_{iter}$: Self-Correction for Predicate Identification}
              \\
            \hline 

            \textbf{User}: For the generated predicate identification result:  \textcolor{llm_text}{predicate identification result.}, rethink the generated predicate identification result, evaluating the accuracy of the output predicate format, verify the correct identification of each predicate, and check for any missing predicates. Identify errors and make corrections accordingly. The predicate identification result format remains consistent with the previous format. 
            \\
            The output format example is: \texttt{"}Issues detected: ...
            Predicate identification result:\texttt{"}. If no errors are found, output \texttt{"}Stop checking.\texttt{"}
            
            \\
            \\
            \textbf{LLM}: \textcolor{llm_text}{Result of self-correction.}

        \end{tabular}
    };
    
    \draw[rounded corners=5pt, line width=1pt] (table.north west) -- (table.north east) -- (table.south east) -- (table.south west) -- cycle;
\end{tikzpicture}
\end{table}

%% file: tables/arg_prompt.tex
\definecolor{process_text}{RGB}{180,0,3}

\definecolor{agent_text}{RGB}{150,10,200}
\definecolor{llm_text}{RGB}{55,180,9}

\begin{table}[t]
    \centering

\begin{tikzpicture}

    \node[inner sep=0pt] (table) {
        \begin{tabular}{p{0.45\textwidth}}
             \rowcolor{gray!15}
            \textbf{Prompt $\mathcal D^{2}_k$: Argument labeling}
              \\
            \hline 
            \dots (\textbf{Prompt} $\mathcal D^{1}$)
            \\
            \\
            \textbf{User}: In SRL, arguments refer to the components or phrases semantically related to a given predicate. They further describe the entities, actions, or concepts associated with the predicate in the sentence.Arguments are divided into core arguments and adjunct arguments.\\
            The labels for all adjunct arguments are as follows:
            \\
            EXT: extent
            \\
            LOC: location
            \\
            \dots 
            \\
            ARGA: secondary agent
            \\
            PRR: predicating relation
            \\
            Core arguments depend on the predicate, and a predicate may have different core argument frames. Within these frames, core arguments will have different interpretations.\\
            Text: \textcolor{process_text}{What @@was\#\# the , purpose and goal of this campaign ?} What are the arguments and their corresponding roles for the given predicate? The predicate is specified by @@ and \#\#.
            \\
            \textcolor{agent_text}{For the predicate \texttt{"}was\texttt{"} in this text, it has the following frames:}
            \\
            \textcolor{agent_text}{For be as a verb: }
            \\
            \textcolor{agent_text}{
            Frame 1: The core arguments it has are: A1: topic, A1: comment.}
            \\
            \textcolor{agent_text}{Frame 2: The core arguments it has are: A1: thing that is.}
            \\
            By referring to the provided frames, determine the frame to which the predicate belongs in order to identify its core arguments. \texttt{"}R-\texttt{"} arguments are arguments that are referencing another argument in the sentence. \texttt{"}C-\texttt{"} arguments are discontinous spans that all refer to the same argument.  \\
             Rewrite the given text and enclose the beginning and end of the arguments with the corresponding <label> and </label> tags.
             \\
            \\
            \textbf{LLM}: \textcolor{llm_text}{Argument labeling result.}
        \end{tabular}
    };
    
    \draw[rounded corners=5pt, line width=1pt] (table.north west) -- (table.north east) -- (table.south east) -- (table.south west) -- cycle;
\end{tikzpicture}
\end{table}

%% file: tables/arg_self_correction.tex
\begin{table}[ht]
    \centering
    
\definecolor{process_text}{RGB}{180,0,3}

\definecolor{agent_text}{RGB}{150,10,200}
\definecolor{llm_text}{RGB}{55,180,9}

\begin{tikzpicture}

    \node[inner sep=0pt] (table) {
        \begin{tabular}{p{0.45\textwidth}}
             \rowcolor{gray!15}
            \textbf{Prompt $C^a_{iter}$: Self-Correction for Argument Labeling}
              \\
            \hline 
            \textbf{User}: For the generated argument labeling result: \textcolor{llm_text}{argument labeling result.},  check the generated argument labeling results for the following issues: whether the generated text is consistent with the original text, and whether the argument label correctly reflects the relationship between the predicate and the argument. Output any identified issues and correct them while maintaining the original argument annotation format. \\
            The output format should follow this example: \texttt{"}Issue detected: ... Argument labeling result: \texttt{"}. If no errors are found, output \texttt{"}Stop checking.\texttt{"}
            \\
            \\
            \textbf{LLM}: \textcolor{llm_text}{Result of self-correction.}

        \end{tabular}
    };
    
    \draw[rounded corners=5pt, line width=1pt] (table.north west) -- (table.north east) -- (table.south east) -- (table.south west) -- cycle;
\end{tikzpicture}
\end{table}

%% file: tables/gpt_template.tex
\begin{table}[t]
    \centering
    
\begin{tikzpicture}

    \node[inner sep=0pt] (table) {
        \begin{tabular}{p{0.45\textwidth}}
             \rowcolor{gray!15}
            \textbf{Prompt: Generate Predicate Explanations}
              \\
               \hline 
              \textbf{User}: You are an expert in summarizing predicate meanings within a given frame. Your task is to infer and summarize the meaning of a predicate based on the provided Chinese predicate and frame arguments. AN represents argument labels in semantic role labeling tasks.  \\

            Here are some examples:  \\
            
            Predicate: abolish  \\
            Frame: A0: entity getting rid of, outlawing something; A1: thing abolished  \\
            Predicate meaning: get rid of, make illegal  \\
            
            Predicate: act  \\
            Frame: A0: agent; A1: predicate  \\
            Predicate meaning: play a role; behave  \\
            
            Predicate: act  \\
            Frame: A0: actor; A1: rounds for action  \\
            Predicate meaning: do something  \\
            
            Predicate: act  \\
            Frame: A0: actor, performer; A1: role, scenario enacted  \\
            Predicate meaning: perform a role  \\
            
            Predicate: \textcolor{process_text}{predicate}  \\
            Frame: \textcolor{process_text}{frame}  \\
            Predicate meaning:  \\
            Directly output the predicate meaning.
           \\
           \\
           \textbf{LLM}: \textcolor{llm_text}{The meaning of the given predicate.}
        \end{tabular}
    };
    
    \draw[rounded corners=5pt, line width=1pt] (table.north west) -- (table.north east) -- (table.south east) -- (table.south west) -- cycle;
\end{tikzpicture}
\end{table}